\documentclass{article}

\pdfoutput=1
\usepackage{arxiv}
\usepackage{graphicx}
\usepackage{booktabs}
\usepackage{multirow}
\usepackage{floatrow}
\usepackage{bbding}
\usepackage{xcolor}
\usepackage{makecell}
\usepackage{amsmath}
\usepackage{soul, color, xcolor}

\usepackage{natbib}
\setcitestyle{numbers,square}
\usepackage[ruled,vlined]{algorithm2e}

\usepackage[utf8]{inputenc} 
\usepackage[T1]{fontenc}    
\usepackage{hyperref}       
\usepackage{url}            
\usepackage{booktabs}       
\usepackage{amsfonts}       
\usepackage{nicefrac}       
\usepackage{microtype}      
\usepackage{xcolor}         
\usepackage{cleveref}
\usepackage{ulem}

\title{Taming Generative Diffusion Prior for Universal Blind Image Restoration}

\author{Siwei Tu, Weidong Yang{$^{\dagger}$}, Ben Fei$^{\dagger}$\\
	Fudan University \\
    \texttt{24110240079@m.fudan.edu.cn}, \texttt{wdyang@fudan.edu.cn}, \texttt{bfei21@m.fudan.edu.cn}\\
}

\date{}

\begin{document}

\maketitle

\begin{abstract}
  Diffusion models have been widely utilized for image restoration. However, previous blind image restoration methods still need to assume the type of degradation model while leaving the parameters to be optimized, limiting their real-world applications. Therefore, we aim to tame generative diffusion prior for universal blind image restoration dubbed \textbf{BIR-D}, which utilizes an \textbf{optimizable convolutional kernel} to simulate the degradation model and dynamically update the parameters of the kernel in the diffusion steps, enabling it to achieve blind image restoration results even in various complex situations. Besides, based on mathematical reasoning, we have provided an empirical formula for the chosen of \textbf{adaptive guidance scale}, eliminating the need for a grid search for the optimal parameter. Experimentally, Our BIR-D has demonstrated superior practicality and versatility than off-the-shelf unsupervised methods across various tasks both on real-world and synthetic datasets, qualitatively and quantitatively. BIR-D is able to fulfill multi-guidance blind image restoration. Moreover, BIR-D can also restore images that undergo multiple and complicated degradations, demonstrating the practical applications.
\end{abstract}

\section{Introduction}
\label{sec:intro}

\begin{figure}[ht]
    \centering
    \includegraphics[width=\linewidth]{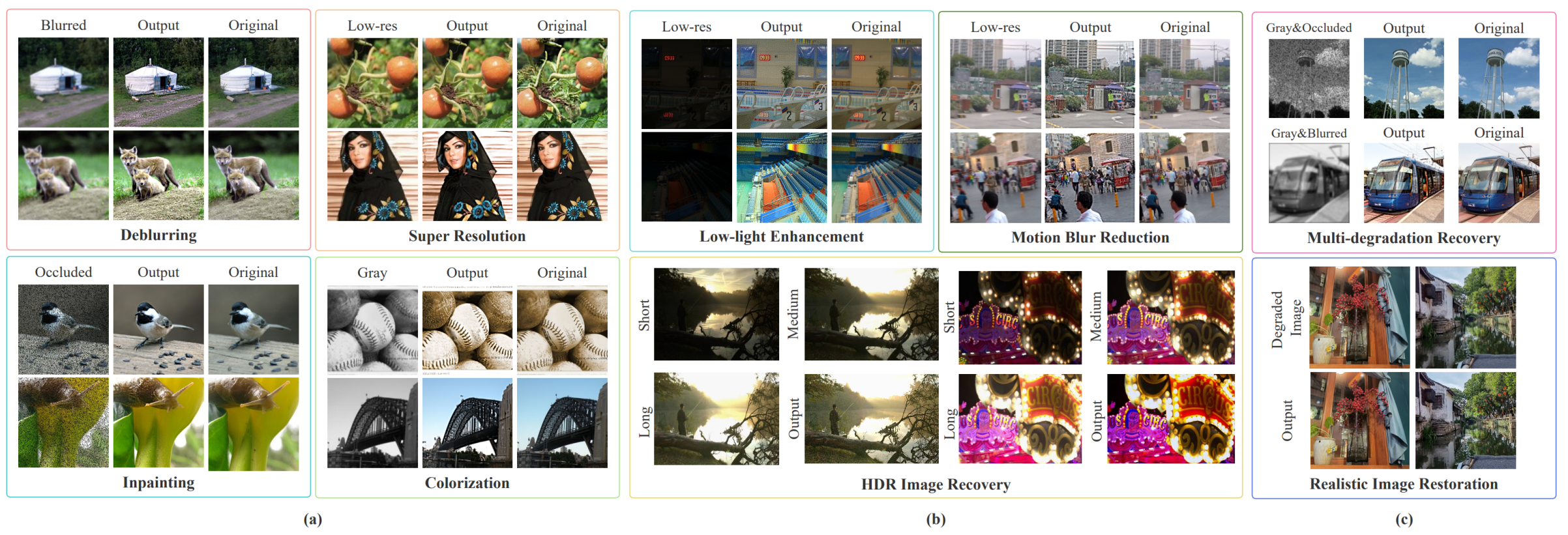}
    \vspace{-0.6cm}
        \caption{Blind Image Restoration Diffusion Model (BIR-D) can achieve high-quality restoration for different types of degraded images. BIR-D not only has the capability to restore (a) linear inverse problems when the degradation function is known. BIR-D can also achieve high-quality image restoration in (b) blind issues with unknown degradation functions, as well as in (c) mixed degradation and real degradation scenarios.}
    \label{mainfig}
\end{figure}

Images inevitably suffer from a degradation in quality in the process of capturing, storing, and compressing. 
Thus, the image restoration task intends to establish a mapping between the degraded image and the original image, to recover a high-quality image from the degraded image. 
In an ideal scenario, the ultimate goal is to undo and restore the degradation process of the image. 
However, in reality, the complexity of the degradation mode often leads to the incapability to fully restore the original high-quality image, which also makes traditional supervised approaches unsuitable for all types of image restoration tasks. 
According to the degradation mode, image restoration tasks can be divided into two types: \textbf{non-blind} and \textbf{blind} problems. 
\textbf{Blind} problems, such as low light enhancement, motion blur reduction and HDR image restoration, refer to image restoration problems where the degradation functions and parameters are totally unknown. 

The blind image restoration problem has attracted increasing attention with the development of generative models. 
The unsupervised blind image restoration methods represented by Generative Adversarial Networks (GANs) \cite{yang2021gan,sophia2022efficient,deng2023hpg,luo2022blind} have the capability to train networks on large datasets of clean images and learn real-world knowledge. 
However, GANs are still difficult to avoid falling into limitations such as poor diversity and difficulty in model training. 
In parallel, diffusion model \cite{wang2023dr2,yi2023diff,laroche2024fast,wu2017color,zhu2023diffusion} have shown strong performance in terms of quality and diversity compared to GANs. 
Pioneer works such as GDP~\cite{fei2023generative}, DDRM~\cite{kawar2022denoising}, and DDNM~\cite{wang2023unlimited} attempt to solve such problems by incorporating the degraded image $y$ as guidance in the sampling process of diffusion models.
By modeling posterior distributions in an unsupervised sampling manner, these approaches showcase the potential for practical guidance in blind image restoration, offering promising implications for real-world applications.
However, the degradation types in these models still need to be assumed, limiting the practicality of natural image restoration where the complicated degradation models always remain unknown.

To this end, we propose an effective and versatile Blind Image Restoration Diffusion Model (BIR-D). 
It utilizes well-trained DDPM~\cite{dhariwal2021diffusion} as an effective prior and is guided by degraded images to form a universal method for various image restoration and enhancement tasks. 
To uniformly model the unknown degradation function of blind image restoration, an optimizable convolutional kernel is dynamically optimized and utilized to simulate the degradation function at each denoising step. 
Specifically, BIR-D updates the convolution kernel parameters based on the gradient of distance loss between the generated image undergoing our optimizable convolutional kernel and the given degraded image. 
At the same time, all existing image restoration methods~\cite{fei2023generative,kawar2022denoising,wang2023unlimited} that use conditional diffusion models manually set the guidance scale as a hyperparameter to control the magnitude of guided generation, which also remains unchanged throughout the sampling process. 
However, for images from different tasks, the guidance scale required for each diffusion step is not entirely the same. 
To deal with this issue, we have derived an empirical formula for the guidance scale, which can calculate the optimal guidance scale for the next denoising step in real-time during the sampling process. 
This improvement avoids the need to manually grid search the optimal value of the guidance scale when solving different tasks and also enhances the quality of generated images. 
With the help of a well-trained DDPM, the above designs enable BIR-D to tackle various blind image restoration tasks.
BIR-D can also achieve multi-degradation or multi-guidance image restoration. 
Furthermore, it showcases satisfactory results in addressing restoration issues related to complex degradation types encountered in real-world scenarios. 

\section{Related Work}
\label{sec:related}

\textbf{Diffusion Model for Image Restoration.}
Image restoration and denoising have seen various advancements with diffusion-based models~\cite{yang2024pgdiff,lin2023diffbir}.
They have been thoroughly explored for linear inverse problems~\cite{kawar2022denoising,fei2023generative}, nonlinear inverse problems~\cite{kawar2022jpeg,fei2023generative}.
To alleviate the fixed- and small-size generation of diffusion models, patch-based algorithm~\cite{ozdenizci2023restoring} and large-size generation~\cite{luo2023refusion,wang2023unlimited} are proposed.
Our model introduces the guidance of degraded images to form an unconditional diffusion model, and attempts to simulate and update the degradation function in real-time, making it suitable for general tasks while maintaining both image quality and efficiency.

\textbf{Blind Image Restoration.}
Many problem-solving approaches have emerged in the field of blind image restoration~\cite{wang2018training}. 
The emergence of GANs~\cite{sophia2022efficient,anand2020tackling} provides several solutions for unsupervised learning in blind image restoration.
On top of GANs, DDPMs are more studied for this task due to the enhanced diversity.
For instance, both DiffBIR~\cite{lin2023diffbir} and GDP~\cite{fei2023generative} leverage generative diffusion priors for blind image restoration. 
BlindDPS~\cite{chung2023parallel} introduces parallel diffusion models for solving blind inverse problems when the functional forms are known. 
PromptIR~\cite{potlapalli2023promptir} uses prompts to encode degradation-specific information and dynamically guide the recovery of the network.
Nevertheless, these methods are still limited to specific tasks. 
BIR-D can be regarded as a unified solver for multiple restoration tasks by simultaneously estimating the recovered images and specific degradation models.

\section{Preliminary}
\label{preliminary}
Diffusion models consists of the forward and reverse processes. 
The forward process continuously adds noise to a natural image $x_0$ through $T$ diffusion steps to obtain the noise distribution $x_T\sim \mathcal{N}(0,I)$, where $\mathcal{N}$ represents the Gaussian distribution. 
The reverse process aims to simulate the noise in each diffusion step and eliminate it, ultimately obtaining the restored generated image $x_0$.

The forward process is a Markov chain defined by the following equation:
\begin{align}
q(x_1,\cdots ,x_T|x_0)=\prod_{t=1}^{T} {q(x_t|x_{t-1})}
\end{align}

It corrupts the initial data $x_0$ into distribution $x_T$ that is close to Gaussian noise after $T$ steps of diffusion, with each sample process defined by $ q (x_t|x_{t-1})=\mathcal{N}(x_t;\sqrt{1-\beta_t}x_{t-1},\beta_tI)$, where $\beta_t$ is the variance of a forward process. 
The variance can be set as a constant or learned by reparameterization. 
Simultaneously defining $\alpha_t=1-\beta_t$,$ \bar{\alpha}_t = {\textstyle \prod_{i=1}^{t}} \alpha_i$. 
It has been proven by \cite{ho2020denoising} that through mathematical reasoning, $x_t$ at any diffusion step can be directly calculated from the starting $x_0$:
\begin{align}
x_t=\sqrt{\bar{\alpha}_t}x_0+\sqrt{1-\bar{\alpha}_t}\epsilon,
\end{align}
where $ \epsilon\sim \mathcal{N}(0,I)$. 
When $T$ is large enough, $\sqrt{\bar{\alpha}_t}$ approaches 0, and at this point $q (x_T | x_0)$ is closer to the latent distribution of $x_T$.

The reverse process is also a Markov chain, which gradually denoises a standard multivariate Gaussian distribution into a denoised image $x_0$. 
Firstly, sample $x_t\sim \mathcal{N}(0, I)$. 
The conditional distribution of the reverse process is $ p_\theta(x_{t-1}|x_t)=\mathcal{N}(x_{t-1};\mu_\theta(x_t,t),\Sigma_\theta I)$. 
According to the Bayesian formula, it can be transformed as follows:
\begin{align}
q(x_{t-1}|x_t,x_0)=q(x_t|x_{t-1},x_0)\frac{q({x_{t-1}|x_0})}{q(x_t|x_0)}
\end{align}

Expand and simplify the three terms at the right end of the equation. 
The variance $\Sigma_\theta$ of the reverse process can be obtained as a fixed value. 
Note that \cite{nichol2021improved} indicates that it can also be learned parameters. 
And the mean of the reverse process $\mu_\theta$ related to $x_t$ and $\tilde{x}_0$:
\begin{align}
\tilde{\mu}_t(x_t,\tilde{x}_0)=\frac{\sqrt{\bar{\alpha}_t-1}\beta_{t}}{1-\bar{\alpha}_t}\tilde{x}_0+\frac{\bar{\alpha}_t(1-\bar{\alpha}_{t-1})}{1-\bar{\alpha}_t}x_t\label{3.1}
\end{align}

According to the formula of the forward process, $\tilde{x}_0$ can be predicted by $x_t$, where $\epsilon$ is a noise function approximator obtained by a neural network $\theta$.
\begin{align}
\tilde{x}_0=\frac{x_t}{\sqrt{\bar{\alpha}_t}}-\frac{\sqrt{1-\bar{\alpha}_t}\epsilon_\theta(x_t,t)}{\sqrt{\bar{\alpha}_t}}
\end{align}

Substitute it into \cref{3.1} to obtain the mean value $\mu_\theta$:
\begin{align}
\mu_\theta(x_t,t)=\frac{1}{\sqrt{\alpha_t}}(x_t-\frac{\beta_t}{\sqrt{1-\bar{\alpha}_t}}\epsilon_\theta(x_t,t))
\end{align}

\section{BIR-D: Universal Blind Image Restoration Diffusion Model}

\begin{figure}[t]
    \centering
    \includegraphics[width=\linewidth]
    {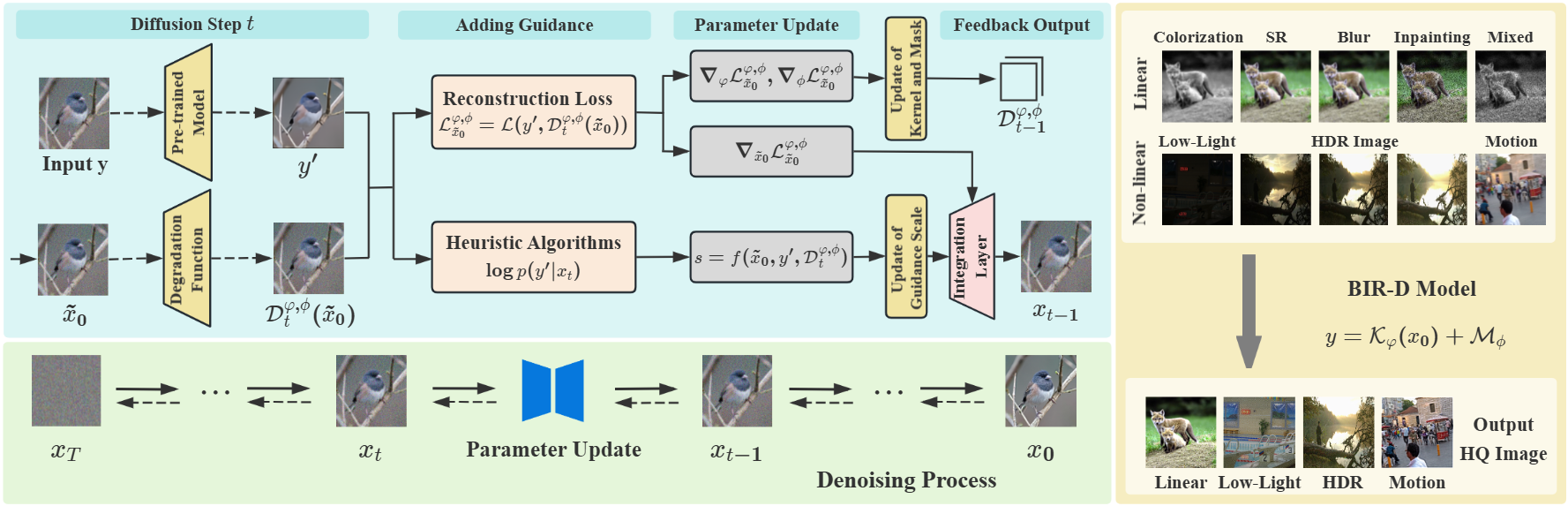}
    \vspace{-0.6cm}
    \caption{\textbf{Overview of BIR-D.} Degraded image $y$ was given during the sampling process. BIR-D systematically incorporates guidance from degraded images in the reverse process of the diffusion model and optimizes the degraded model at the same time. For degraded image $y$, pre-training is first performed to provide a better initial state for BIR-D. 
    BIR-D introduces a distance function in each step of the reverse process of the diffusion model to describe the distance loss between the degraded image $y$ and the generated image $\tilde{x}_0$ after the degradation function, so that the gradient could be used to update and simulate a better degradation function. Based on the empirical formula, the adaptive guidance scale can be calculated to provide optimal guidance during the sampling process.}
    \label{fig.flowchart}
\end{figure}

In this study, we aim to use a well-trained DDPM~\cite{dhariwal2021diffusion} to learn the prior distribution of images and ultimately solve non-blind and blind problems in various image restoration tasks.

\subsection{Optimizable convolutional kernel as a universal degradation function}
For a natural image $x$, its corresponding degraded image $y$ can be obtained by the degradation function $y=\mathcal{D}(x)$. 
Most of the blind image restoration methods~\cite{fei2023generative,wang2023unlimited} are used to solve the situation where the degradation function $\mathcal{D}$ is known while leaving the parameters of $\mathcal{D}$ are unknown. 
However, when dealing with real-world image restoration problems, the degradation function $\mathcal{D}$ is not only an unknown quantity but also difficult to accurately represent mathematically. 
Therefore, we propose an optimized convolutional kernel to simulate complex degradation functions. 
The parameters of the convolution kernel in the degradation function are dynamically optimized along with the denoising steps.

Moreover, in the real-world scenario, considering that there are different noises in different subtle areas of the image, using only one optimized convolutional kernel may not fully cover this situation.
Therefore, we propose to utilize a mask $\mathcal{M}$ to model and estimate these noises. 
Thus, the entire degradation process can be represented as: $y=K(x)+\mathcal{M}$, where $\mathcal{K}$ refers to the optimized convolutional kernel used in the model and $\mathcal{M}$ is a mask with the same dimension as image $x$. 
$\mathcal{K}$ and $\mathcal{M}$ have their own optimizable parameters, forming the degradation function $\\mathcal{D}$. 
In this way, any degradation process can be simulated by this degradation function.

\begin{algorithm}[b]\small
\caption{Conditional diffusion model with the guidance of degraded image $y$, given a diffusion model noise prediction function
$\epsilon_\theta(x_t,t)$.}
\label{alg.1}
\KwIn{Degraded image $y$, degradation function $\mathcal{D}$ composed of optimized convolutional kernels $\mathcal{K}$ with parameters $\varphi$ and mask $\mathcal{M}$ with parameters $\phi$, learning rate $l$, distant measure $\mathcal{L}$. }
\KwOut{Output image $x_0$ conditioned on $y$.}
Sample $x_T$ from $\mathcal{N}(0,I)$

    \For{t from T to 1}{
        $\tilde{x} _0=\frac{x_t}{\sqrt{\bar{\alpha}_t }}-\frac{\sqrt{1-\bar{\alpha}_t}\epsilon_\theta(x_t,t)}{\sqrt{\bar{\alpha}_t }}$\
        
        $\mathcal{L}_{\varphi,\phi,\tilde{x}_0} =\mathcal{L}(y,\mathcal{D}^{\varphi,\phi}(\tilde{x}_0))$\
        
        $s = -\frac{(x_t-\mu)^Tg+C+\log_{}{N}}{\mathcal{L}(\mathcal{D}^{\varphi,\phi}(\tilde{x_0}),y)}$\

        $\tilde{x}_0 \gets \tilde{x}_0-\frac{s(1-\bar{\alpha}_t) }{\sqrt{\bar{\alpha}_{t-1}}\beta_t}\nabla_{{\tilde{x}}_0}\mathcal{L}_{\varphi,\phi,\tilde{x}_0}$\
        
       $\tilde{\mu}_t=\frac{\sqrt{\bar{\alpha}_{t-1}}\beta_t}{1-\bar{\alpha}_t}\tilde{x}_0+\frac{\sqrt{\bar{\alpha}_{t}}(1-\bar{\alpha}_{t-1})}{1-\bar{\alpha}_t}{x}_t$\

        $\tilde{\beta}_t=\frac{1-\bar{\alpha}_{t-1}}{1-\bar{\alpha}_t}\beta_t$\
        
        Sample $x_{t-1}$ from $\mathcal{N}(\tilde{\mu}_t,\tilde{\beta}_tI)$\

        $\varphi \gets \varphi-l\nabla_{\varphi}\mathcal{L}_{\varphi,\phi,\tilde{x}_0}$\

        $\phi \gets \phi-l\nabla_{\phi}\mathcal{L}_{\varphi,\phi,\tilde{x}_0}$\

    }
\textbf{return} $x_0$
\end{algorithm}

\begin{figure}[t]
    \centering
    \includegraphics[width=\linewidth]{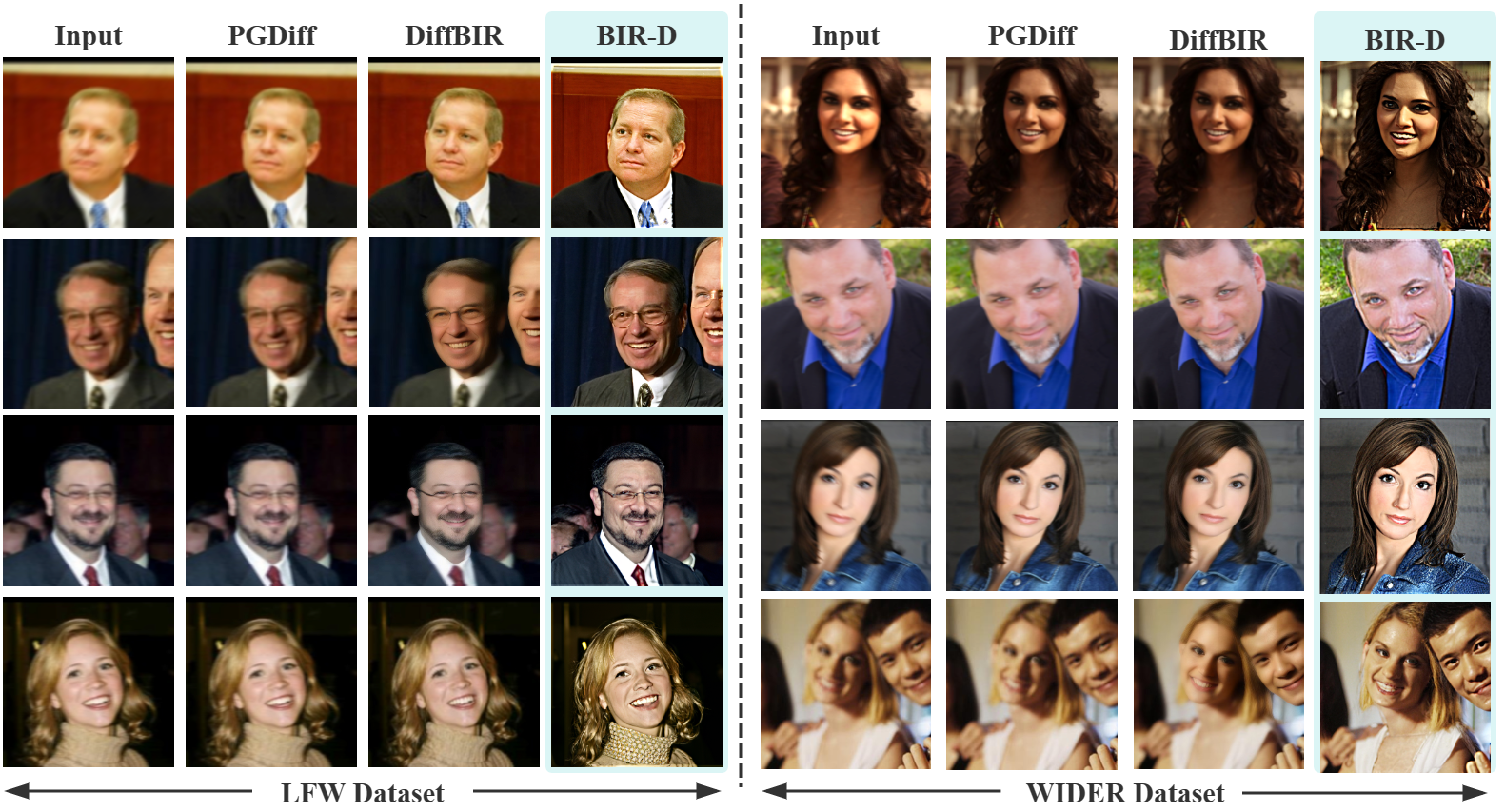}
    \vspace{-0.7cm}
    \caption{Comparison of image quality for blind face restoration results on LFW \cite{wang2021towards} and WIDER dataset \cite{zhou2022towards}. }
    \label{real_world_fig}
\end{figure}

\begin{table}[t]\small
  \centering
  \caption{\textbf{Quantitative comparison of blind face restoration on LFW and WIDER datasets}}
  \vspace{-0.3cm}
  \label{real_world_tab}
  \begin{tabular}{c|c c|c c}
    \toprule[1pt]
     \multirow{2}{*}{\qquad\textbf{Task} \qquad } & \multicolumn{2}{c|}{\textbf{LFW dataset}}& \multicolumn{2}{c}{\textbf{WIDER dataset}}\\
     \cmidrule(lr){2-5}
     
    & \; FID  & \; NIQE & \; FID  & \; NIQE  \\
    \midrule
    PGDiff~\cite{yang2024pgdiff} & \;71.62 & \; 4.15 & \; 39.17 & \; 3.93 \\
    DiffBIR~\cite{lin2023diffbir} & \; \textbf{39.58} & \; 4.03 & \; 32.35 & \; 3.78 \\
    \midrule
    BIR-D & \; 40.12 & \; \textbf{3.94} & \; \textbf{31.49} & \; \textbf{3.65} \\
    \bottomrule[1pt]
  \end{tabular}
\vspace{-0.3cm}
\end{table}

\begin{table}[t]\small
  \centering
  \caption{\textbf{Quantitative comparison of linear inverse problems on ImageNet 1k\cite{pan2021exploiting}. }}
  \setlength\tabcolsep{0.6pt}
  \vspace{-0.3cm}
  \label{Linear problems}
  \resizebox{\textwidth}{!}{
  \begin{tabular}{c|c c c c |c c c c | c c c c | c c c c}
    \toprule[1pt]
     \multirow{2}{*}{\textbf{Task}}&\multicolumn{4}{c|}{\textbf{4$\times$Super resolution}}&\multicolumn{4}{c|}{\textbf{Deblur}}&\multicolumn{4}{c|}{\textbf{25$\%$ Inpainting}}& \multicolumn{4}{c}{\textbf{Colorization}}\\
     \cmidrule(lr){2-17}
     
    &PSNR&SSIM&Consistency&FID&PSNR&SSIM&Consistency&FID&PSNR&SSIM&Consistency&FID&PSNR&SSIM&Consistency&FID\\
    \midrule
    RED\cite{romano2017little}&24.18&0.71 &27.57&98.30&21.30&0.58&63.20&69.55&- &-&-&-&-&-&-&-\\
    DGP\cite{pan2021exploiting}&21.65&0.56&158.74&152.85&26.00&0.54&475.10&136.53&27.59&0.82&414.60&60.65&18.42&0.71&305.59&94.59\\
    SNIPS\cite{kawar2021snips}&22.38&0.66&21.38&154.43&24.73&0.69&60.11&17.11&17.55&0.74&587.90&103.50&-&-&-&-\\
    DDRM\cite{kawar2022denoising}&\textbf{26.53}&0.78&19.39&40.75&\textbf{35.64}&\textbf{0.98}&50.24&4.78&34.28&0.95&\textbf{4.08}&24.09&\textbf{22.12}&0.91&37.33&47.05\\
    DDNM\cite{wang2022zero}&25.36&\textbf{0.81}&7.52&39.14&24.66&0.71&41.70&4.64&32.16&\textbf{0.96}&5.42&17.63&21.95&0.89&36.41&38.79\\
    GDP\cite{fei2023generative}&24.42&0.68&6.49&38.24&25.98&0.75&41.27&2.44&\textbf{34.40}&\textbf{0.96}&5.29&16.58&21.41&\textbf{0.92}&36.92&37.60\\
    \midrule
    BIR-D&24.58&0.71&\textbf{6.32}&\textbf{37.54}&26.31&0.73&\textbf{38.42}&\textbf{2.32}&33.59&0.90&5.18&\textbf{15.73}&22.09&0.89&\textbf{36.12}&\textbf{36.58}\\
    \bottomrule[1pt]

  \end{tabular}
  }
\vspace{-0.3cm}
\end{table}

\begin{figure}[t]
    \centering
    \includegraphics[width=\linewidth]{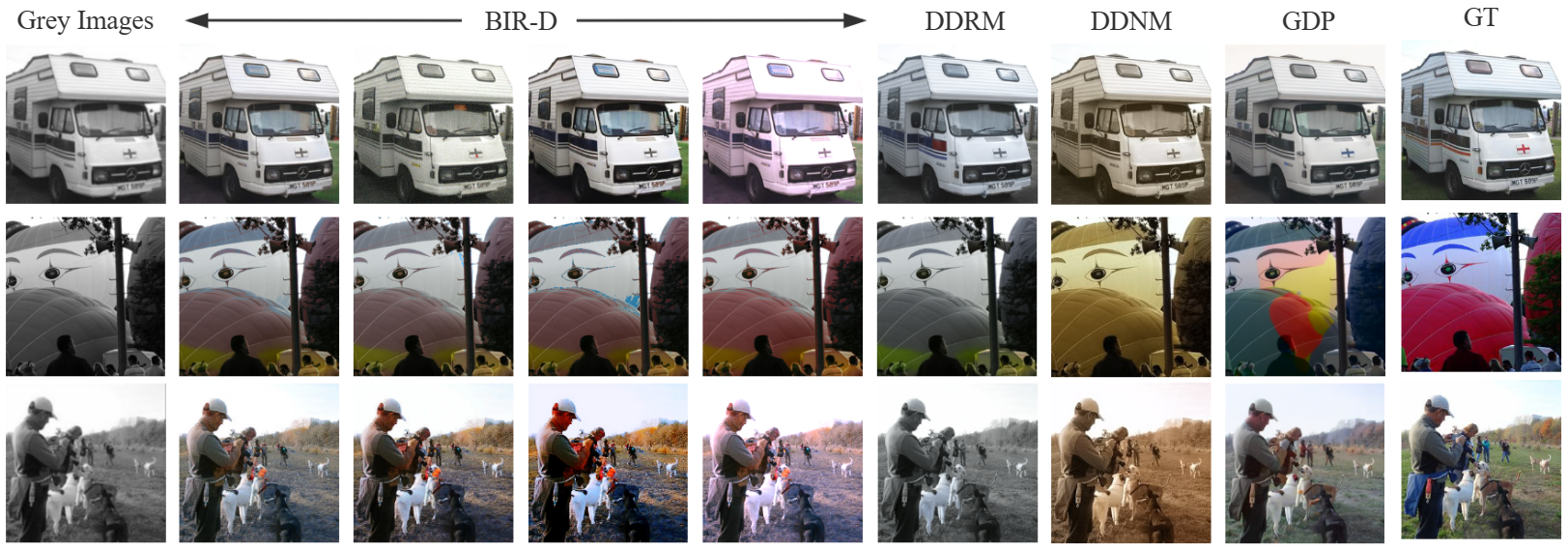}
    \vspace{-0.7cm}
    \caption{\textbf{Comparison of colorization image on ImageNet 1k\cite{pan2021exploiting}.} BIR-D can generate various outputs on the same input image.}
    \label{col}
\end{figure}

\begin{figure}[t]
    \centering
    \includegraphics[width=\linewidth]{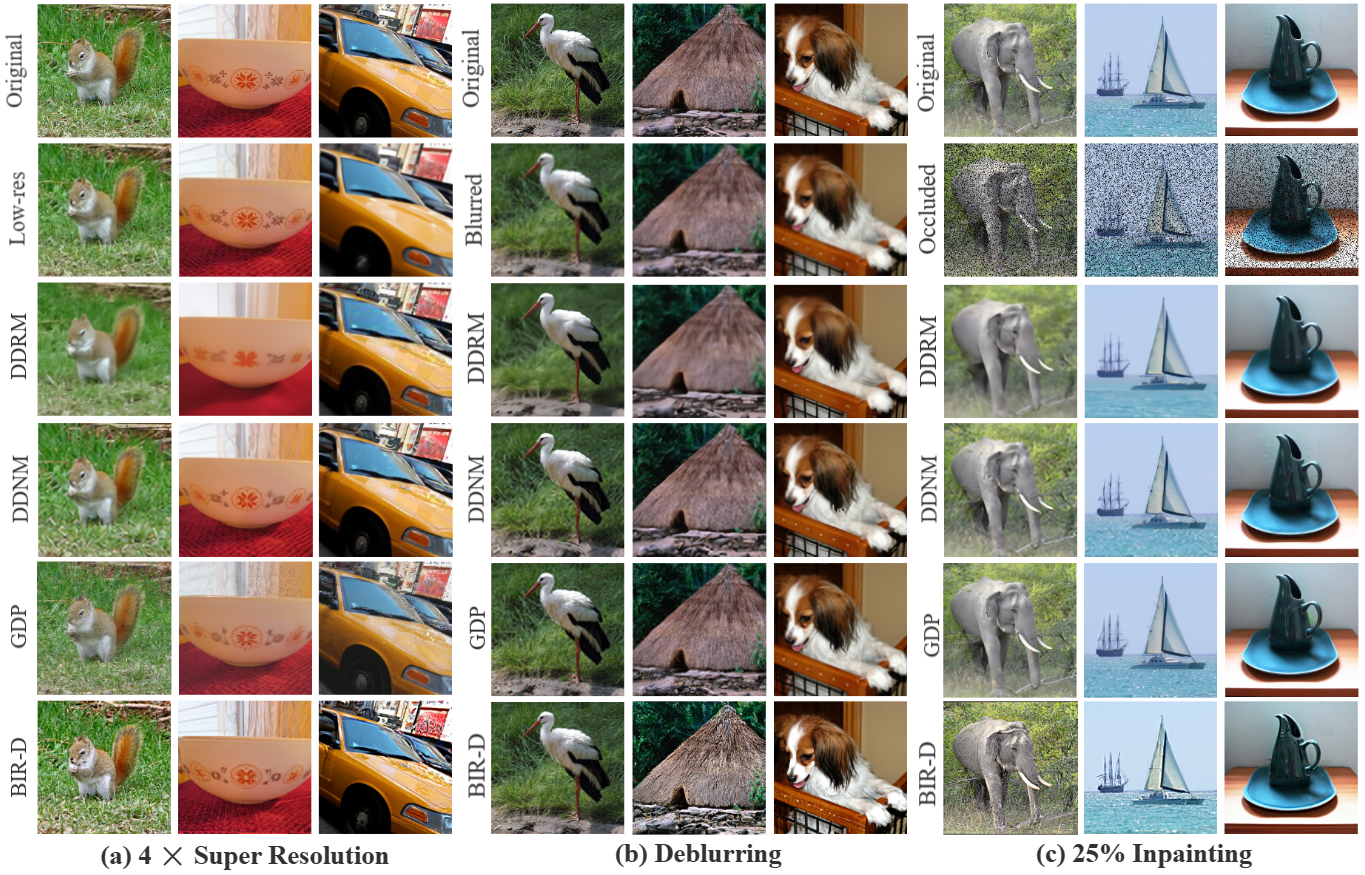}
    \vspace{-0.6cm}
    \caption{Results of linear degradation tasks on 256 × 256 images from ImageNet 1k.}
    \label{linear problems figure}
\end{figure}

\begin{figure}[t]
    \centering
    \includegraphics[width=\linewidth]{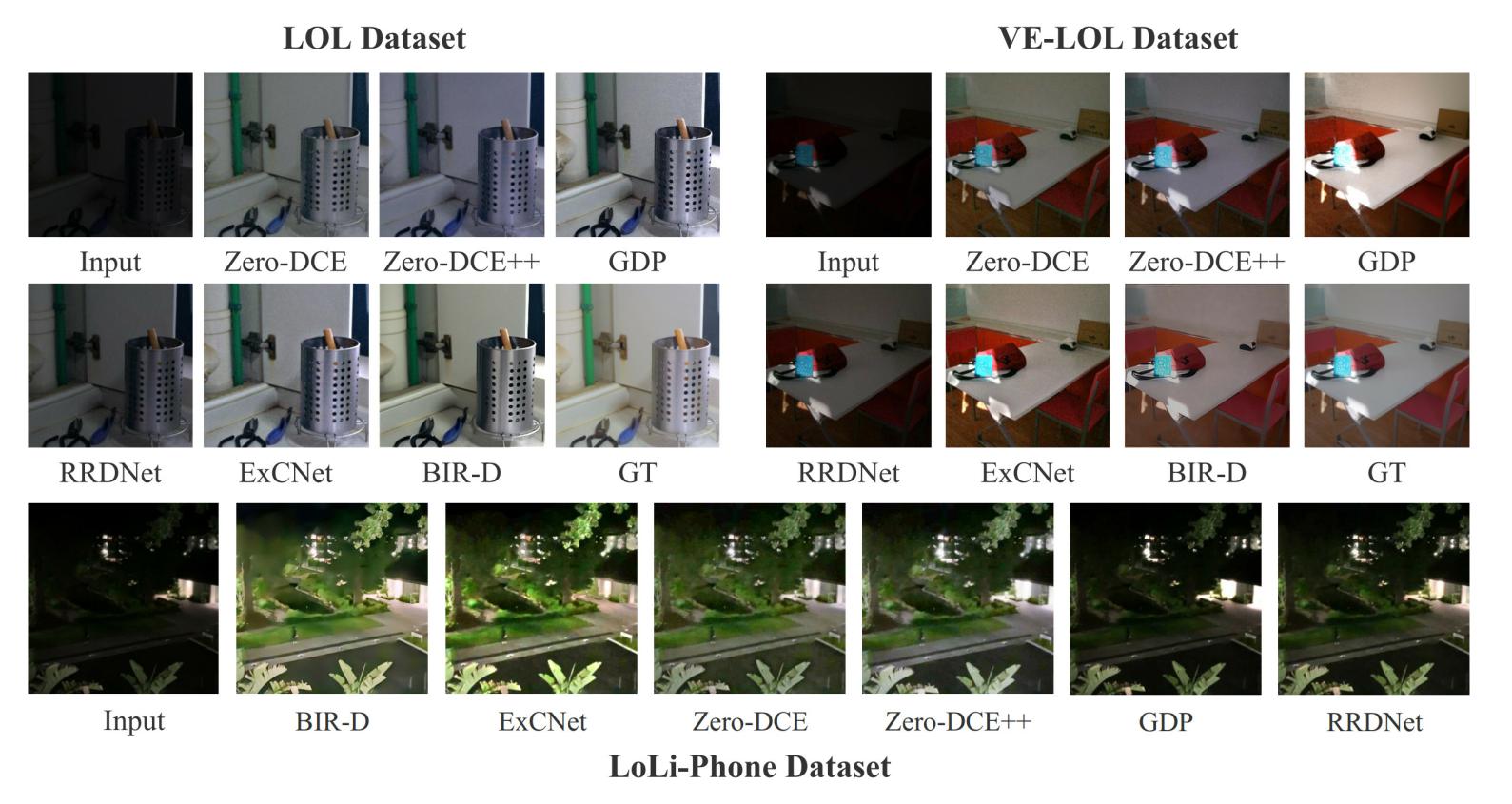}
    \vspace{-0.8cm}
    \caption{Comparison of image quality in low-light enhancement task on the LoL \cite{wei1808deep}, VE-LOL \cite{liu2021benchmarking} and LoLi-Phone \cite{li2021low} datasets.}
    \label{Low-Light}
\end{figure}

\begin{table}[t]\small
  \centering
  \caption{\textbf{Quantitative comparison among various zero-shot learning methods of low-light enhancement task on LOL \cite{wei1808deep} and VE-LOL-L \cite{liu2021benchmarking}} Bold font represents the best metric result. }
  \vspace{-0.2cm}
  \label{tab_lowlight}
  \resizebox{\textwidth}{!}{
  \begin{tabular}{c|c c c c c |c c c c c}
    \toprule[1pt]
     \multirow{2}{*}{\;\textbf{Task} \; } & \multicolumn{5}{c|}{\textbf{LOL}}& \multicolumn{5}{c}{\textbf{VE-LOL-L}}\\
     \cmidrule(lr){2-11}
     
    & \; PSNR  & \; SSIM  & LOE &  FID  &  PI &\;  PSNR  & \; SSIM  &  LOE &  FID  & PI    \\
    \midrule
    ExCNet\cite{zhang2019zero} & \;\textbf{16.04} & \;0.62 & 220.38 & 111.18 & 8.70 & \;16.20 & \;0.66 & 225.15 & 115.24 & 8.62 \\
    Zero-DCE\cite{guo2020zero} & \;14.91 & \;\textbf{0.70} & 245.54 & 81.11 & 8.84 & \;\textbf{17.84} & \;\textbf{0.73} & 194.10 & 85.72 & 8.12 \\
    Zero-DCE$++$\cite{li2021learning} & \;14.86 & \;0.62 & 302.06 & 86.22 & 7.08  & \;16.12 & \;0.45 & 313.50 & 86.96 & 7.92\\
    RRDNet\cite{zhu2020zero} & \;11.37 & \;0.53 & 127.22 & 89.09 & 8.17  & \;13.99 & \;0.58 & 94.23 & 83.41 & 7.36\\
    GDP\cite{fei2023generative} & \;13.93 & \;0.63 & 110.39 & 75.16 & 6.47  & \;13.04 & \;0.55 & 79.08 & 78.74 & 6.47\\
    \midrule
    BIR-D & \;14.52 & \;0.56 & \textbf{105.42} & \textbf{68.98} & \textbf{4.87} & \;13.87 & \;0.51 & \textbf{78.18} & \textbf{74.54} & \textbf{5.73}\\
    \bottomrule[1pt]

  \end{tabular}
  }
\vspace{-0.3cm}
\end{table}

\begin{figure}[t]
    \centering
    \includegraphics[width=\linewidth]{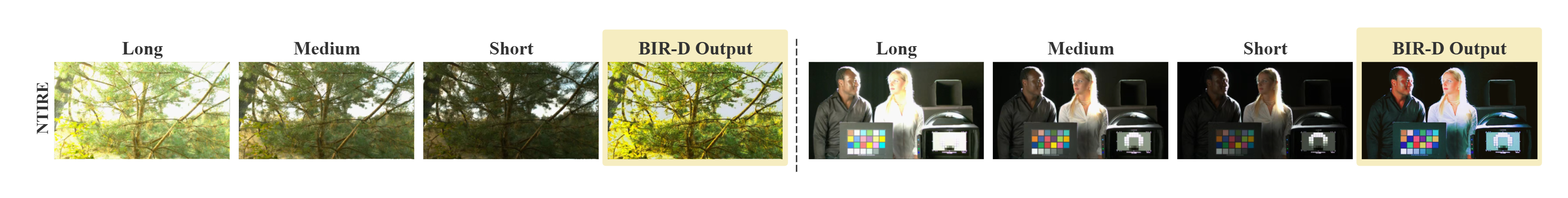}
    \vspace{-0.8cm}
    \caption{Comparison of image quality for HDR image recovery results on NTIRE \cite{perez2021ntire}. }
    \label{HDR_3}
\end{figure}

\begin{figure}[t]
    \centering
    \includegraphics[width=\linewidth]{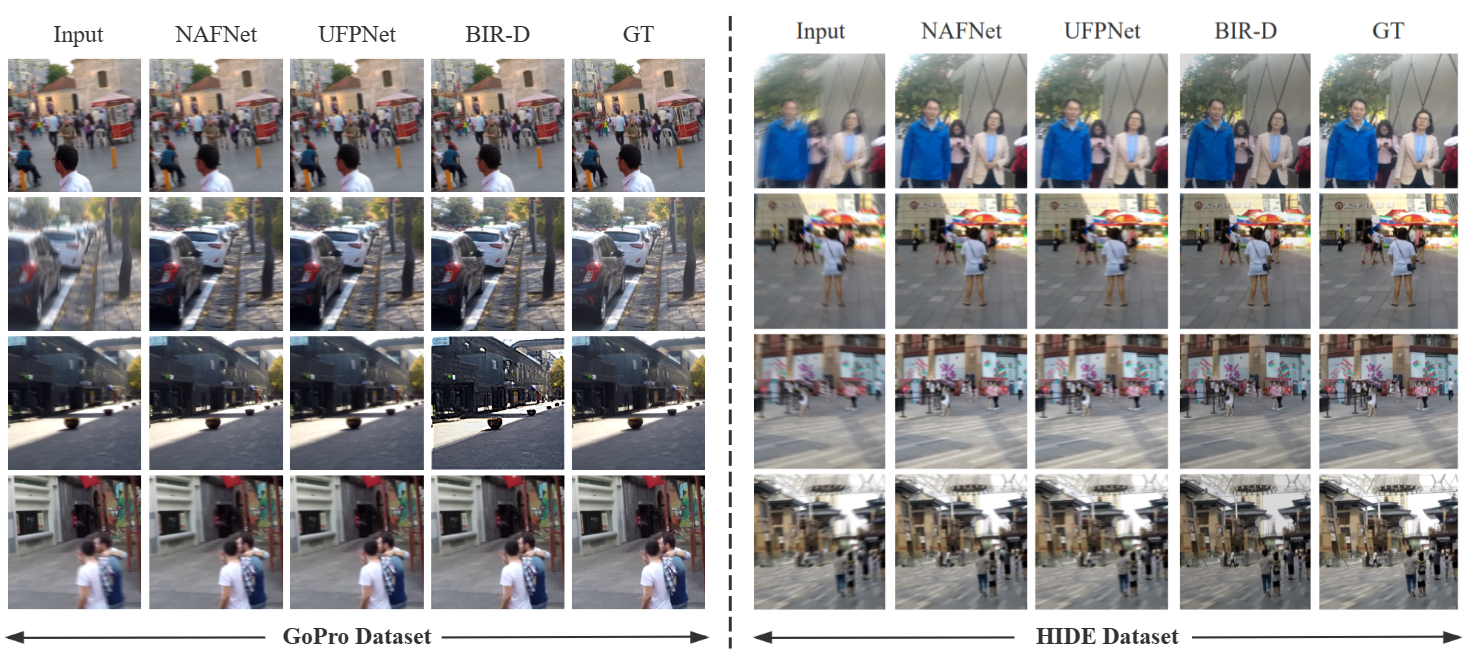}
    \vspace{-0.7cm}
    \caption{Comparison of image quality for motion blur reduction results on GoPro \cite{nah2017deep} and HIDE dataset \cite{shen2019human}. }
    \label{motionblurfig}
\end{figure}

\begin{table*}[t]\small
\centering
\caption{\textbf{Quantitative comparison of motion blur reduction and HDR image recovery tasks. }}
\vspace{-0.3cm}
\resizebox{\textwidth}{!}{
\begin{tabular}{c|c c|c c |c | c c c c}
    \toprule[1pt]
     \multirow{2}{*}{\textbf{Motion Blur Reduction
     }} &\multicolumn{2}{c|}{\textbf{GoPro}} & \multicolumn{2}{c|}{\textbf{HIDE}}& \multirow{2}{*}{\textbf{HDR Recovery
     }}&\multicolumn{4}{c}{\textbf{NTIRE}}\\
     \cmidrule(lr){2-5}
     \cmidrule(lr){7-10}

    & PSNR& SSIM    &   PSNR    &  SSIM   &&   PSNR    &  SSIM & LPIPS & FID\\
    \midrule
    DeepRFT\cite{mao2023intriguing} & 33.23 & 0.963 & 31.42 & 0.944 &Deep-HDR\cite{wu2018deep} &  21.66 &  0.76 &  0.26 &  57.52 \\
    MSDI-Net\cite{li2022learning}& 33.28 & 0.964 & 31.02 & 0.940 &AHDRNet\cite{yan2019attention}&  18.72 &  0.58 &  0.39 &  81.98\\
    NAFNet\cite{chen2022simple}& 33.69 & 0.967 & 31.32 & 0.943 &HDR-GAN\cite{niu2021hdr}& 21.67 & 0.74 & 0.26 & 52.71\\
     UFPNet\cite{fang2023self} & 34.06 & \textbf{0.968} & 31.74 & 0.947 &GDP\cite{fei2023generative} &  24.88 &  0.86 &  \textbf{0.13} &  50.05\\
    \midrule 
    BIR-D& \textbf{34.12} & \textbf{0.968} & \textbf{32.09} & \textbf{0.948} &BIR-D & \textbf{25.03} & \textbf{0.88} & 0.16 & \textbf{48.74}\\
    \bottomrule[1pt]

  \end{tabular}
  }
 \label{Tab3}
\vspace{-0.3cm}
\end{table*}

\begin{figure}[t]\small
    \centering
    \includegraphics[width=\linewidth]
    {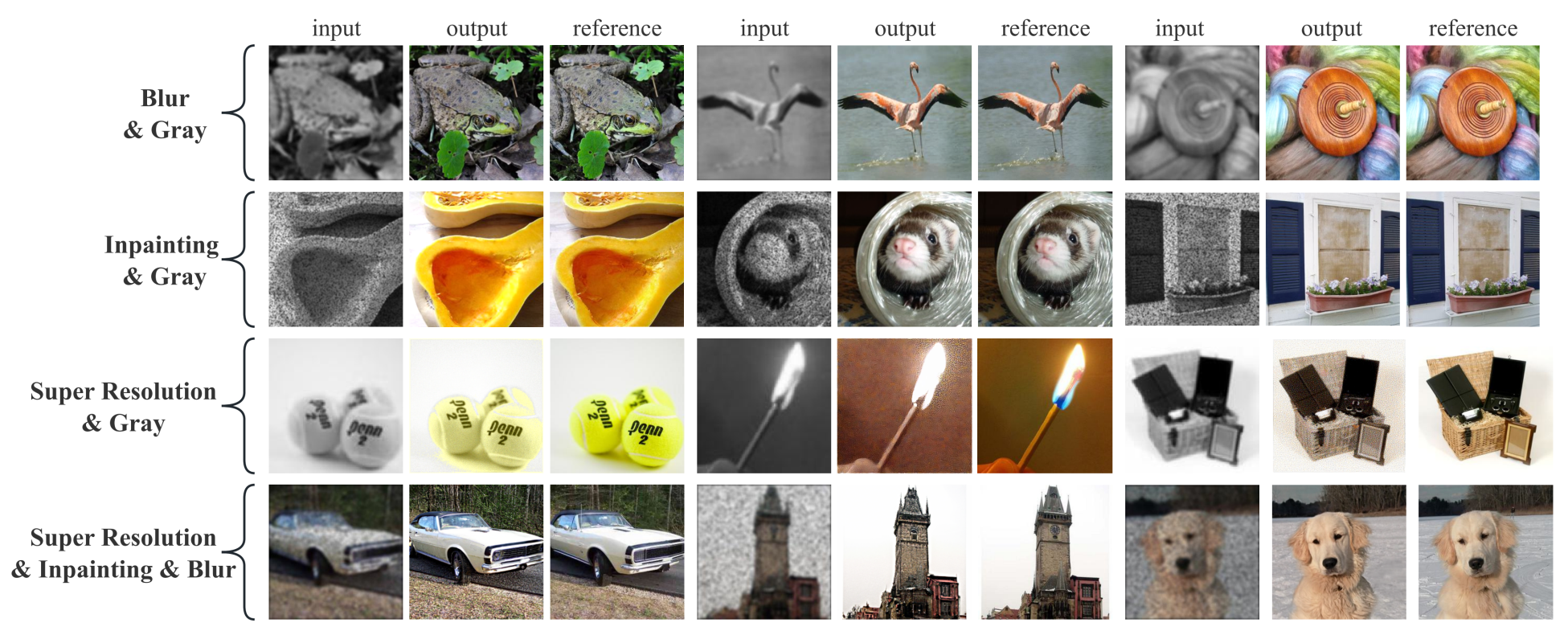}
    \vspace{-0.7cm}
    \caption{Results of multi-task image restoration.}
    \label{fig.multi}
\end{figure}

\begin{table}[b]\small
  \centering
  \caption{\textbf{The ablation study on the optimizable convolutional kernel and the empirical settings of guidance scale.}}
  \vspace{-0.3cm}
  \label{k and s}
  \begin{tabular}{c|c c|c c c c c | c c}
    \toprule[1pt]
     \multirow{2}{*}{\textbf{{\small Methods}}} &\multicolumn{2}{c}{\textbf{{\small Dynamic Update}}}&\multicolumn{5}{|c|}{\textbf{{\small LOL}}}&\multicolumn{2}{c}{\textbf{{\small LoLi-Phone}}}\\
     \cmidrule(lr){2-3}
     \cmidrule(lr){4-10}

    &{\small Kernel}&{\small Guidance Scale}&{\small PSNR}&{\small SSIM}&{\small LOE}&{\small FID}&{\small PI}&{\small LOE}&{\small PI}\\
    \midrule
    {\small Model A}&\XSolidBrush&\XSolidBrush&8.96&0.46&210.88&113.36&8.24&110.05& 8.36\\
    {\small Model B}&\XSolidBrush&\Checkmark&9.58&0.48&203.83&102.47&7.90&102.55&8.25\\
    {\small Model C}&\Checkmark&\XSolidBrush&14.35&0.54&113.56&82.14&5.23&75.34&7.94\\
    \midrule
    {\small BIR-D}&\Checkmark&\Checkmark&\textbf{14.52}&\textbf{0.56}&\textbf{105.42}&\textbf{68.98}&\textbf{4.87}&\textbf{72.83}&\textbf{6.12}\\
    \bottomrule[1pt]

  \end{tabular}
\vspace{-0.3cm}
\end{table}

\begin{table}[t]
  \centering
  \caption{\textbf{The ablation study of kernel size in blind issues.}}
  \vspace{-0.3cm}
  \label{kernel size}
  \begin{tabular}{c|c c c c c |c c}
    \toprule[1pt]
     \multirow{2}{*}{\qquad\textbf{Task} \qquad } & \multicolumn{5}{c|}{\textbf{Low-Light Enhancement}}& \multicolumn{2}{c}{\textbf{Motion Blur Reduction}}\\
     \cmidrule(lr){2-8}
     
    & \; PSNR  & \; SSIM  & \; LOE & \; FID  & \; PI & \; \qquad PSNR  & \; SSIM  \\
    \midrule
    kernel size$=$1 & \;13.73 & \;0.49 & \;118.38 & \;78.52 & \;5.67 & \;\qquad 31.14 & \; 0.917 \\
    kernel size$=$3 & \;13.90 & \;0.54 & \;113.89 & \;74.41 & \;5.24 & \;\qquad 32.07 & \; 0.937 \\
    kernel size$=$7 & \;14.47 & \;\textbf{0.56} & \;108.75 & \;70.55 & \;4.93 & \;\qquad 33.94 & \; 0.961 \\
    \midrule
    \makecell[c]{BIR-D with\\5 $\times$ 5 kernel} & \;\textbf{14.52} & \;\textbf{0.56} & \;\textbf{105.42} & \;\textbf{68.98} & \;\textbf{4.87} & \;\qquad \textbf{34.12} & \; \textbf{0.968} \\
    \bottomrule[1pt]
  \end{tabular}
\vspace{-0.3cm}
\end{table}
\begin{table}[t]
  \centering
  \caption{\textbf{The ablation study of kernel size in linear inverse problem.}}
  \vspace{-0.3cm}
  \label{linear kernel size}
  \resizebox{\textwidth}{!}{
  \begin{tabular}{c|c c c c |c c c c}
    \toprule[1pt]
     \multirow{2}{*}{\qquad\textbf{Task} \qquad } & \multicolumn{4}{c|}{\textbf{4 $\times$ Super resolution}}& \multicolumn{4}{c}{\textbf{Deblur}}\\
     \cmidrule(lr){2-9}
     
    & \; PSNR  & \; SSIM  & \;Consistency & \; FID  & \; PSNR  & \; SSIM  & \; Consistency & \; FID  \\
    \midrule
    kernel size=13 & \; 24.05 & \; 0.66 & \; 6.65 & \; 39.02 & \; 26.12 & \; 0.74 & \; 41.29 & \; 3.09 \\
    kernel size=7 & \; 24.31 & \; 0.67 & \; 6.64 & \; 38.91 & \; 26.53 & \; 0.77 & \; 38.60 & \; 2.53 \\
    kernel size=11 & \; 24.36 & \; 0.69 & \; 6.50 & \; 38.07 & \; 26.79 & \; 0.79 & \; 38.52 & \; 2.44 \\
    \midrule
    \makecell[c]{BIR-D with\\9 $\times$ 9 kernel} & \; \textbf{24.58} & \; \textbf{0.71} & \; \textbf{6.32} & \; \textbf{37.54} & \; \textbf{27.14} & \; \textbf{0.84} & \; \textbf{37.86} & \; \textbf{2.32} \\
    \bottomrule[1pt]
        
     \multirow{2}{*}{\qquad\textbf{Task} \qquad } & \multicolumn{4}{c|}{\textbf{25$\%$ Inpainting}}& \multicolumn{4}{c}{\textbf{Colorization}}\\
     \cmidrule(lr){2-9}
     
    & \; PSNR  & \; SSIM  & \; Consistency & \; FID & \; PSNR  & \; SSIM  & \; Consistency & \; FID    \\
    \midrule
    kernel size=7 & \; 29.58 & \; 0.80 & \; 6.17 & \; 18.09 & \; 20.07 & \; 0.76 & \; 39.85 & \; 42.29 \\
    kernel size=13 & \; 31.12 & \; 0.84 & \; 5.64 & \; 16.56 & \; 21.04 & \; 0.83 & \; 37.71 & \; 38.14  \\
    kernel size=11 & \; 32.91 & \; 0.86 & \; 5.41 & \; 16.17 & \; 21.57 & \; 0.85 & \; 37.69 & \; 38.01 \\
    \midrule
    \makecell[c]{BIR-D with\\9 $\times$ 9 kernel} & \; \textbf{33.59} & \; \textbf{0.90} & \; \textbf{5.18} & \; \textbf{15.73} & \; \textbf{22.09} & \; \textbf{0.89} & \; \textbf{36.12} & \; \textbf{36.58} \\
    \bottomrule[1pt]

  \end{tabular}
  }
\vspace{-0.3cm}
\end{table}

\begin{table}[t]\small
  \centering
  \caption{\textbf{The ablation study on the effectiveness of the pre-training model.}}
  \label{pre-training model}
  \vspace{-0.2cm}
  \resizebox{\textwidth}{!}{
  \begin{tabular}{c|c c c c | c c c c}
    \toprule[1pt]
     \multirow{2}{*}{\textbf{Task} } & \multicolumn{4}{c|}{\textbf{Random initial value}}  & \multicolumn{4}{c}{\textbf{Biased initial value}}\\
     \cmidrule(lr){2-9}

    & \;PSNR  & SSIM   & Consistency  & FID & \; PSNR  &  SSIM  &  Consistency &  FID     \\
    \hline
    \makecell[c]{ BIR-D without\\pre-training model}
  & \;25.88 & 0.69 & 40.24 & 2.55 & 21.49  & 0.61 & 53.78 & 4.32 \\
    \hline
    BIR-D & \;26.31 & 0.73 & 38.42 & 2.32 & 25.97  & 0.71 & 39.87 & 2.41\\
    \bottomrule[1pt]

  \end{tabular}
}
\vspace{-0.3cm}
\end{table}

\subsection{Empirical formula of guidance scale}
In the reverse denoising process of DDPM, the generated images can be conditioned on degraded image $y$~\cite{batzolis2021conditional}. 
Specifically, the distribution $p_\theta({x_{t-1}|x_t})$ of reverse denoising is converted into a conditional distribution $ p_\theta(x_{t-1}|x_t,y)$. 
It is demonstrated~\cite{dhariwal2021diffusion} that the difference between it and the original formula lies in the addition distribution of $ p (y|x_t)$, which serves as a probability representation for denoising $x_t$ into a high-quality image consistent with $y$. 
Previous work \cite{fei2023generative} proposed a feasible calculation to approximate this indicator by using heuristic algorithms:
\begin{align}
\log_{}{ p(y\mid x_{t})}=- \log_{}{N}-s\mathcal{L}( \mathcal{D}(\tilde{x}_0),y))\label{4.1.1},
\end{align}
where $N$ is the normalization factor, which is the distribution $ p_\theta(y|x_{t+1})$, and $s$ is the scalar factor used to control the importance of guidance, named guidance scale. 
$\mathcal{L}$ is the distance metric. 

The value of the guidance scale plays a crucial role in the quality of the image generation result. 
A larger value can lead to overall blurring of the image, while a smaller value can result in missing details in the restoration.
However, the guidance scale in existing works~\cite{fei2023generative,zhu2023diffusion,wang2023unlimited} can only be manually set as a hyperparameter. 
But in specific experiments, the optimal value of the guidance scale varies in different masks, degraded images, and diffusion steps. 
The original configuration necessitates thorough testing for the initial setup. Additionally, employing the same guidance scale for every denoising step is not an optimal choice.

Therefore, we propose an empirical formula for the guidance scale, which can dynamically calculate and update the optimal values of guidance factors in real-time at each diffusion step of degraded images in specific repair tasks.
Specifically, we noticed that the distribution $ \log_{}{p_\theta(y|x_t)} $ can be applied to perform Taylor expansion around $x=\mu$ and take the first two terms. 

\begin{align}
\log_{}{ p_\theta(y\mid x_{t})}&\approx \log_{}{ p(y\mid x_t)}\mid_{x_t=\mu}+(x_t-\mu)^T\nabla_{x_t}\log_{}{ p_\theta(y\mid x_t)}\mid_{x_t=\mu} \\
&=(x_t-\mu)^Tg+C,
\end{align}
where $g=\nabla_{x_t}\log_{}{ p_\theta(y\mid x_t})\mid_{x_t=\mu}$, $C=\log_{}{ p(y\mid x_t)}\mid_{x_t=\mu}$. 
By combining the heuristic approximation formula and Taylor expansion formula mentioned above, we can simplify the empirical formula for the guidance scale:

\begin{align}
\label{gs}
s = -\frac{(x_t-\mu)^Tg+C+\log_{}{N}}{\mathcal{L}(\mathcal{D}(x_t),y)}
\end{align}

For each image at every moment $t$, the applicable value of the guidance scale can be calculated.
However, because $y$ is a degraded image without noise, while $x_t$ itself has noise. The use of MSE loss between $x_t$ and $y$ can lead to the introduction of noise into the guidance process. Therefore, we utilize the MSE error between the estimated value $\tilde{x}_0$ and $y$ here, and the above formula could to be corrected as:

\begin{align}
s = -\frac{(x_t-\mu)^Tg+C+\log_{}{N}}{\mathcal{L}(\mathcal{D}(\tilde{x}_0),y)}.
\end{align}
The guidance scale is related to the generated images $x$, degraded image $y$, and the degradation function $\mathcal{D}$. 
This value of this \textbf{Adaptive Guidance Scale} can be dynamically updated in each diffusion step so that each step in the diffusion model can use the most appropriate guidance scale. 

\subsection{Sampling process of BIR-D}

Through empirical formulas, we can obtain the conditional transition formula in the reverse process of the diffusion model.
\begin{align}
\log_{}{p_\theta(x_{t}|x_{t+1},y)}&=\log_{}{(p_\theta(x_{t}|x_{t+1})p(y|x_{t}) }+N_1\label{4.2.1} \\
&\approx \log_{}{p(z)+N_2},\label{4.2.2}
\end{align}
where $z$ conforms to the distribution $ \mathcal{N}(z;\mu_\theta(x_t,t)+\Sigma g,\Sigma)$. 
The intermediate quantity $ g=\nabla_{x_t}\log_{}{p(y|x_t)}$.
The value of $g$ can be obtained by calculating the gradient in heuristic algorithms in \cref{4.1.1}, which includes the parameter of guidance scale:
\begin{align}
g=\nabla _{x_t}\log_{}{p(y|x_t)}=-s\nabla_{x_t}\mathcal{L}(\mathcal{D}(x_t),y)
\end{align}
The other terms ${N_1},{N_2}$, and the variance of the reverse process $\Sigma= \Sigma_\theta (x_t)$ in \cref{4.2.1} and \cref{4.2.2} are constants, and the unconditional distribution $p_\theta (x_ {t-1} | x_t)$ is given by traditional diffusion models.

Therefore, the conditional transition distribution $p (x_{t-1} | x_t, y)$ can be approximately estimated by adding $ -(s\Sigma\nabla_{x_t}\mathcal{L}(\mathcal{D}(x_t),y))$ to the mean of the traditional unconditional transition distribution.
Previous studies~\cite{fei2023generative} have shown that the addition of $\Sigma$ has a negative impact on the quality of generated images.
Therefore, in this experiment, we omitted the term $\Sigma$, and the complete sampling process is shown in \cref{alg.1}.

Detailly, in the diffusion step $t$ of the sampling process, the noise of $x_t$ is first predicted from the given pre-trained DDPM and eliminated to obtain an estimated value of $x_0$. 
Subsequently, apply the degradation function of step $t$ to $x_0$ and calculate its reconstruction loss with the degraded image $y$. 
We utilize our adaptive guidance scale for sampling the next step latent $x_{t-1}$. 
In this process, it is necessary to calculate the gradient about $x_0$ and the parameters of each convolution kernel in the distance metric loss, which is used to update the convolution kernel parameters in real time for the next sampling process.

\textbf{Pre-process.}
The empirical formula for the guidance scale we construct is related to the degradation function.
Herein, when the model simulates the degradation function more reasonably, BIR-D can obtain more appropriate guidance scale values accordingly.
To this end, we introduce a first-stage pre-training model from~\cite{lin2023diffbir} to further enhance the model's capability to correct initial deviations.
This enables the model to have a strong correction ability for significant deviations in the degradation function during the initial diffusion step, ultimately generating ideal image restoration results.

\textbf{Multi-degradation Image Restoration.}
In the real world, the degradation process often involves a combination of multiple different complex types. 
To improve the image restoration capability of the model in complex situations and enhance its practicality, we propose to extend BIR-D into multi-task scenarios. 
To our surprise, BIR-D can fulfill multi-degradation image restoration without any modification (\Cref{fig.multi}) thanks to the mixture of degradation types can also be simulated as an unknown degradation by an optimizable convolutional kernel.  

\section{Experiments}
In this section, we systematically compare BIR-D with other blind image restoration methods in real-world and synthetic datasets.

\textbf{Blind Image Restoration on Real-world Datasets.}
Firstly, we evaluate the blind image restoration capability of BIR-D on two real-world datasets, namely LFW dataset~\cite{wang2021towards} and WIDER dataset~\cite{zhou2022towards}. 
As shown in \Cref{real_world_fig}, BIR-D successfully simulated and removed blur, and achieved more ideal facial detail restoration.
The quantitative results in \Cref{real_world_tab} shows that BIR-D outperforms PGDiff~\cite{yang2024pgdiff} and DiffBIR~\cite{lin2023diffbir} in NIQE metric on both datasets and FID metric on WIDER, demonstrating better blind image restoration performance.

\textbf{Comparison on Common Linear Inverse Problems.}
We conducted experiments on linear inverse problems on ImageNet 1k to compare BIR-D with off-the-shelf methods.
For each experiment, we calculated the average Peak Signal-to-Noise Ratio (PSNR), Structural Similarity (SSIM), Consistency, and FID results, where PSNR, SSIM, and Consistency are used to quantify the faithfulness between the generated image and the original image, while FID is used to measure the quality of the generated image. 
To make fair comparisons, other methods are given known degradation functions as reported in the original paper while BIR-D utilizes universal degradation functions for different tasks.
\Cref{Linear problems} shows that BIR-D outperforms other methods in terms of Consistency and FID in almost all tasks.
As shown in \Cref{linear problems figure}, the images generated by BIR-D demonstrate a high level of image quality and details.
Moreover, \Cref{col} also demonstrates that BIR-D can generate various results in image restoration tasks.

\textbf{Low Light Enhancement.}
We further evaluated the effectiveness of BIR-D in low-light image enhancement. 
Following the previous works~\cite{fei2023generative}, we utilized three datasets, LOL \cite{wei1808deep}, VE-LOL-L \cite{liu2021benchmarking}, and LoLi-Phone~\cite{li2021low}, to test the restoration ability of BIR-D. 
As shown in \Cref{tab_lowlight}, our BIR-D outperforms all the zero-shot methods in both FID and Lightness Order Error (LOE) \cite{wang2013naturalness}, and demonstrates significant improvement in Perceptual index (PI) \cite{mittal2012making}. 
A lower PI value reflects better perceptual quality, while a lower LOE reflects a better natural preservation ability of the generated image, making images to have a more natural sensory experience. 
As shown in \Cref{Low-Light}, BIR-D exhibits reasonable and well-exposed results.

\textbf{HDR Image Recovery.}
In the HDR image restoration task, we compared BIR-D with other leading methods, including DeepHDR \cite{wu2018deep}, AHDRNet \cite{yan2019attention}, HDR-GAN \cite{niu2021hdr}, and GDP \cite{fei2023generative}, on the NTIRE2021 Multi-Frame HDR Challenge \cite{perez2021ntire} dataset. 
The quantitative and qualitative results are presented in \Cref{Tab3} and \Cref{HDR_3}, with BIR-D showing the best PSNR and SSIM levels, and successfully generating results with rich and accurate detailed information.

\textbf{Motion Blur Reduction.}
To evaluate the performance of BIR-D in the motion blur reduction tasks, we compare BIR-D with the state-of-the-art motion blur reduction methods on GoPro dataset \cite{nah2017deep} and HIDE dataset \cite{shen2019human}. 
We used the same input image, which also means that the motion blur of the input image is the same, ensuring fairness in comparison. 
The comparison results of the metrics are presented in \Cref{Tab3}, where BIR-D outperforms existing methods in both PSNR and SSIM. 
As shown in \Cref{motionblurfig}, BIR-D can effectively achieve the elimination of motion blur. 
The generated images not only achieve a better quality but also receive restoration with more clear details.

\textbf{Multi-Degradation Image Restoration.}
Encouraged by the excellent restoration performance of BIR-D on single restoration task, we further tested the image restoration performance of BIR-D in solving multi-task image restoration. 
As shown in \Cref{fig.multi}, we take a degraded image on the ImageNet dataset where two types of degradation are mixed as an example. 
The optimizable convolution kernel of BIR-D can also simulate these complicated degradation functions. 
The generated images obtained have excellent results in both image quality and details.

\subsection{Ablation study}
\label{Ablation study}

\begin{figure}[t]
    \centering
    \includegraphics[width=\linewidth]
    {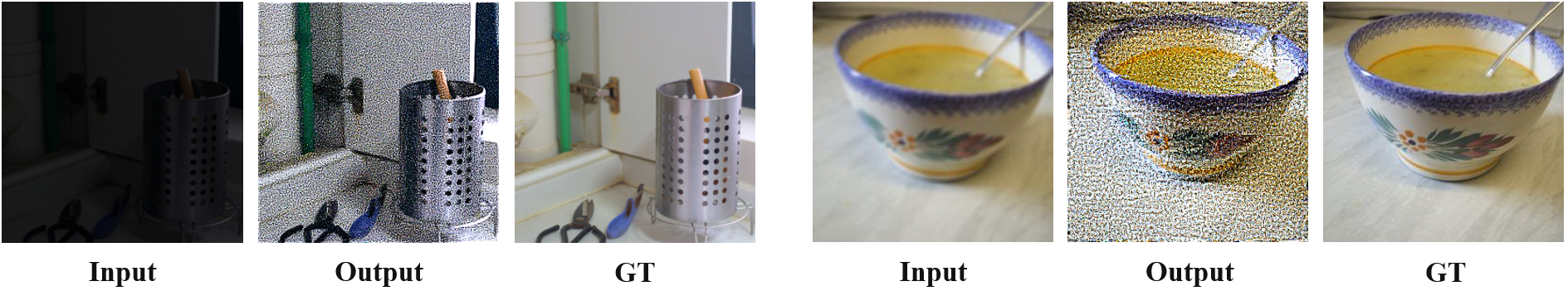}
    \vspace{-0.6cm}
    \caption{Qualitative results when the fixed guidance scale is biased towards a larger value of $s=80000$. }
    \label{fig.Fixed}
\end{figure}

\textbf{The Effectiveness of Optimizable Convolutional Kernel and Adaptive Guidance Scale.}
The ablation studies on the real-time optimizable convolutional kernel parameters and guidance scale were performed to reveal the effectiveness of these settings. 
We further tested the LOL~\cite{wei1808deep} and the most challenging LoLi-phone~\cite{li2021low} datasets. 
Model A fixed the convolutional kernel parameters and guidance scale.
Models B and C represent fixed parameters for the convolution kernel and the fixed guidance scale, respectively.
As shown in \Cref{k and s}, BIR-D outperformed other models in all indicators, demonstrating the effectiveness of an optimizable convolutional kernel and adaptive guidance scale.

\textbf{The Optimal Size of Optimizable Convolutional Kernel.}
\label{kernel}
In the main paper, in order to assure the versatility of BIR-D, we used convolution kernels of size $7\times7$ for all tasks. 
Nevertheless, for different types of tasks, the size of the convolution kernel might be different.
To explore the impact of kernel size on the quality of generated images, we conducted experiments using convolution kernels of different sizes in various types of image restoration tasks.
As shown in \Cref{kernel size}, for blind image restoration tasks, the experiment showed that the results of a 5$\times$5 convolution kernel perform best. 
For linear inverse tasks (\Cref{linear kernel size}), the optimal convolution kernel size was 9$\times$9. 

\textbf{The Effectiveness of the First Stage Pre-training Model.}
We conducted further experiments on the deblur task to demonstrate the impact of the first-state pre-training model. 
As shown in \Cref{pre-training model}, for a randomly initialized convolution kernel parameter, all metrics of BIR-D were better than BIR-D without the pre-training model.
These results indicate that the first-stage pre-trained model is able to provide better initial state of images for our BIR-D.

\section{Parameter analysis}
\label{App:Kernel_2}
\begin{figure}[t]
    \centering
    \includegraphics[width=\linewidth]
    {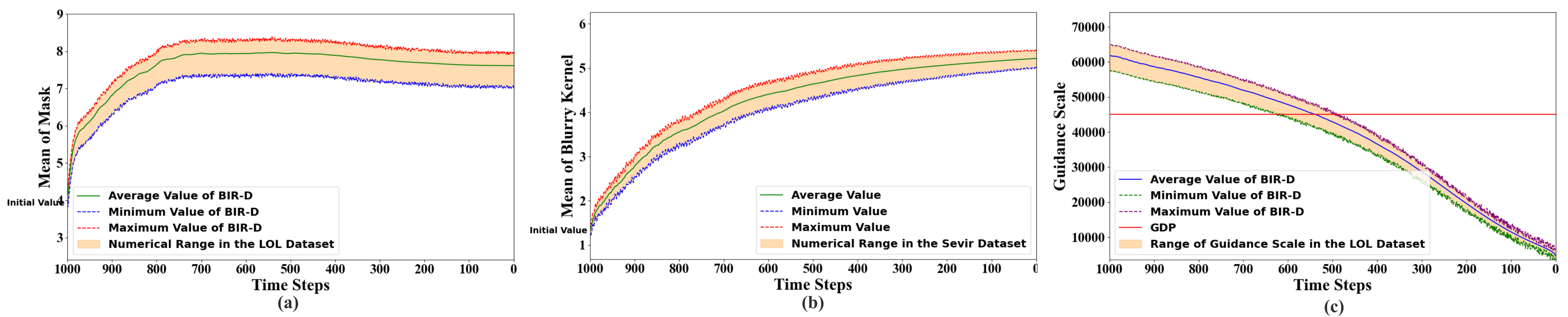}
    \vspace{-0.6cm}
    \caption{Illustration of \textbf{(a)} the variation of the mean of optimizable convolutional kernel parameters in each step of the sampling process. \textbf{(b)} The variation of the mean of degradation mask in each step of the sampling process. \textbf{(c)} The variation of adaptive guidance scale in each step of the sampling process. These experiments are performed on LOL dataset. }
    \label{fig.kernel_mask}
\end{figure}

\textbf{The Parameter Variations of the Optimizable Convolution Kernel and Mask in the Reverse Steps.} 
In order to visualize the variation trends of the parameters of convolution kernel mask in the reverse process, we conducted experiments on the test set of the LOL dataset from the low-light enhancement task. 
As shown in \Cref{fig.kernel_mask}(a), the mean values of the convolution kernel parameters and degradation mask are given by random initialization and gradually increase with the progress of the time steps. 
This increase in magnitude is influenced by the gradient of the distance metric with respect to the corresponding parameters. 
When the sampling step $t$<500, the difference between $\tilde{x}_0$ and $y$ changes slightly, resulting in correspondingly smaller gradient values. 

\begin{figure}[h!]
    \centering
    \includegraphics[width=\linewidth]
    {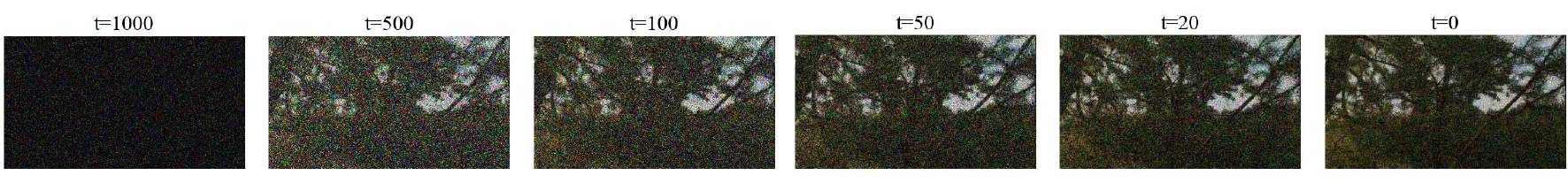}
    \vspace{-0.7cm}
    \caption{The changing of degradation mask during the sampling process in HDR recovery. }
    \label{fig.mask_hdr}
\end{figure}

BIR-D employs masks in the degradation function with the intent to address the image restoration of local regions characterized by substantial shifts in brightness. 
\Cref{fig.kernel_mask}(b) shows that the mask $\mathcal{M}$ of the degradation model has an upward trend from their initial values, making the overall degradation function approach the true degradation. 
As shown in \Cref{fig.mask_hdr}, during the sampling process, the degradation mask learns the detailed information of the image, including local regions with significant brightness differences. 
This process is obtained by updating the gradient of the distance metric with respect to the degradation mask parameters.

\textbf{The Theoretical Analysis of the Changing Trend of Guidance Scale in the Reverse Steps.}
We take the variation in the guidance scale of BIR-D on the LOL dataset as an example to analyze the trend of its changes during the reverse steps.
As shown in \Cref{fig.kernel_mask}(c), the guidance scale gradually decreases with the sampling step, which aligns with the actual situation. 
When the sampling step $t$<500, as $t$ decreases, the difference between $x_t$ and $x_{t-1}$ decreases with decreasing $t$, indicating a reduction in the simulated noise at each step. 
Therefore, the level of guidance required for each sampling step should also be reduced accordingly, leading to a decrease in the required guidance scale values.  
According to \Cref{gs}, when step $t$ is small, the gradient term $g$ also decreases due to the small change in $x_t$ at each step. 
The speed of the gradient term decreases is greater than the speed of distance metric decreases, resulting in a decrease in the value of the guidance scale. 

\section{Conclusion}
\label{sec:conclusion}

In this paper, we propose Blind Image Restoration Diffusion, which is a unified model that can be used to solve various blind image restoration problems. 
We utilize optimized convolutional kernels to simulate and update the degradation function in the diffusion step in real time, and derive the empirical formula of the guidance scale in detail, so that it can better utilize the conditional diffusion model to generate high-quality images. 
The ability to solve various blind image restoration tasks, including low-light enhancement and motion blur reduction, has also been verified through various indicators of datasets. 

\section*{Acknowledgment}
The authors would like to thank Zhaoyang Lyu for his technical assistance. 

\bibliographystyle{unsrt}
\bibliography{main}

\end{document}